\documentclass[11pt]{article}
\usepackage{coling2018}
\usepackage{times}
\usepackage{url}
\usepackage{latexsym}
\usepackage{graphicx}
\usepackage{xcolor}
\usepackage{subcaption}
\usepackage{todonotes}
\usepackage{multirow}

\title{RiTUAL-UH at TRAC 2018 Shared Task: Aggression Identification}

\author{Niloofar Safi Samghabadi \qquad Deepthi Mave \qquad Sudipta Kar \qquad
Thamar Solorio \\
Department of Computer Science\\ University of Houston \\
Houston, TX 77204-3010\\
  {\tt \{nsafisamghabadi, dmave, skar3, tsolorio\}@uh.edu} \\}

\date{}

\begin{document}
\maketitle
\begin{abstract}
This paper presents our system for ``TRAC 2018 Shared Task on Aggression Identification''. Our best systems for the English dataset use a combination of lexical and semantic features. However, for Hindi data using only lexical features gave us the best results. We obtained weighted F1-measures of 0.5921 for the English Facebook task (ranked 12th), 0.5663 for the English Social Media task (ranked 6th), 0.6292 for the Hindi Facebook task (ranked 1st), and 0.4853 for the Hindi Social Media task (ranked 2nd). 
\end{abstract}

\section{Introduction}
\label{intro}

\blfootnote{
    %
    %
    %
    %
    %
    %
    \hspace{-0.65cm}  
    This work is licensed under a Creative Commons 
    Attribution 4.0 International License.
    License details:
    \url{http://creativecommons.org/licenses/by/4.0/}.
}

Users' activities on social media is increasing at a fast rate. Unfortunately, a lot of people misuse these online platforms to harass, threaten, and bully other users. This growing aggression against social media users has caused serious effects on victims, which can even lead them to harm themselves. The TRAC 2018 Shared Task on Aggression Identification \cite{trac2018report} aims at developing a classifier that could make a 3-way classification of a given data instance between ``Overtly Aggressive'', ``Covertly Aggressive'', and ``Non-aggressive''.
We present here the different systems we submitted to the shared task, which mainly use lexical and semantic features to distinguish different levels of aggression over multiple datasets from Facebook and other social media that cover both English and Hindi texts.  

\section{Related Work}

In recent years, several studies have been done towards detecting abusive and hateful language in online texts. Some of these works target different online platforms like Twitter~\cite{waseem2016hateful}, Wikipedia~\cite{Wulczyn:2016}, and ask.fm~\cite{samghabadi2017detecting} to encourage other research groups to contribute to aggression identification in these sources.

Most of the approaches proposed to detect offensive language in social media make use of multiple types of hand-engineered features. ~\newcite{nobata2016abusive} use n-grams, linguistic, syntactic and distributional semantic features to build a hate speech detection framework over Yahoo! Finance and News and get an F-score of 81\% for a combination of all features.~\newcite{davidson2017automated} combine n-grams, POS-colored n-grams, and sentiment lexicon features to detect hate speech on Twitter data.~\newcite{VanHee15detect} use word and character n-grams along with sentiment lexicon features to identify nasty posts in ask.fm.~\newcite{samghabadi2017detecting} build a model based on lexical, semantic, sentiment, and stylistic features to detect nastiness in ask.fm. They also show the robustness of the model by applying it to the dataset from different other sources. 


Based on \newcite{malmasi2018challenges}, distinguishing hate speech from profanity is not a trivial task and requires features that capture deeper information from the comments. In this paper, we try different combinations of lexical, semantic, sentiment, and lexicon-based features to identify various levels of aggression in online texts.




\section{Methodology and Data}

\subsection{Data}

\begin{table}[h!]
\centering

\begin{tabular}{|l|r|r|r|r|}
\hline
\textbf{Data}    & \multicolumn{1}{l|}{\textbf{Training (FB)}} & \multicolumn{1}{l|}{\textbf{Validation (FB)}} & \multicolumn{1}{l|}{\textbf{Test (FB)}} & \multicolumn{1}{l|}{\textbf{Test (SM)}} \\ \hline
\textbf{English} & 12000                                  & 3001                                      & 916                                     & 1257                                    \\ \hline
\textbf{Hindi}   & 12000                                  & 3001                                      & 970                                     & 1194                                    \\ \hline
\end{tabular}
\caption{Data distribution for English and Hindi corpus}
\label{table1}
\end{table}
The datasets were provided by \newcite{trac2018dataset}. Table~\ref{table1} shows the distribution of training, validation and test (Facebook and social media) data for English and Hindi corpora. The data has been labeled with one out of three possible tags:
\begin{itemize}
\item \textbf{Non-aggressive (NAG):} There is no aggression in the text. 
\item \textbf{Overtly aggressive (OAG):} The text is containing either aggressive lexical items or certain syntactic structures.
\item \textbf{Covertly aggressive (CAG):} The text is containing an indirect attack against the target using polite expressions in most cases.
\end{itemize}

\subsection{Data Pre-processing}
Generally the data from social media resources is noisy, grammar and syntactic errors are common, with a lot of ad-hoc spellings, that make it hard to analyze. Therefore, we first put our efforts to clean and prepare the data to feed it to our systems.
For the English dataset, we lowercased the data and removed URLs, Email addresses, and numbers. We also did minor stemming by removing ``ing'', plural and possessive ``s'', and replaced a few common  abstract grammatical forms with the formal versions.

On manual inspection of the training data for Hindi, we found that some of the instances are Hindi-English code-mixed, some use Roman script for Hindi and others are in Devanagari. Only 26\% of the training data is in Devanagari script. We normalize the data by transliterating instances in Devanagari to Roman script. These instances are identified using Unicode pattern matching and are transliterated to Roman script using \textit{indic-trans} transliteration tool\footnote{\url{https://github.com/libindic/indic-trans}}. For further analysis, we run an in-house word-level language identification system on the training data \cite{mave2018lid}. This CRF system is trained on Facebook posts and has an F1-weighted score of 97\%. Approximately 60\% of the training data is code-mixed, 39\% is only Hindi and $0.42\%$ is only English. 

\subsection{Features} \label{sec:feautures}
We make use of the following features:

\noindent\textbf{Lexical: }Words are powerful mediums to convey a feeling, describe or express ideas. With this notion, we use word $n$-grams (n=1, 2, 3), char $n$-grams (n=3, 4, 5), and $k$-skip $n$-grams (k=2, n=2, 3) as features. We weigh each term with its term frequency-inverse document frequency (TF-IDF). We also consider using another weighting scheme by trying binary word $n$-grams (n=1, 2, 3). 
~\\
\noindent\textbf{Word Embeddings: } The idea behind this approach is to use a vector space model for extracting semantic information from the text~\cite{le2014distributed}. For the embedding model we use pre-trained vectors trained on part of Google News dataset including about 3 million words\footnote{\url{https://code.google.com/archive/p/word2vec/}}. We computed word embeddings feature vectors by averaging the word vector of all the words in each comment. We skip the words which are not in the vocabulary of the pre-trained model. This representation is only used for English data and the coverage of the Google word embedding is 63\% for this corpus.

\noindent\textbf{Sentiment: }We use Stanford Sentiment Analysis tool~\cite{socher2013recursive}\footnote{\url{https://nlp.stanford.edu/sentiment/code.html}} to extract fine-grained sentiment distribution of each comment. For every message, we calculate the mean and standard deviation of sentiment distribution over all sentences and use them as feature vector. 

\noindent\textbf{LIWC (Linguistic Inquiry and Word Count): }LIWC2007 \cite{Pennebaker07liwc} includes around 70 word categories to analyze different language dimensions. In our approach, we only use the categories related to positive or negative emotions and self-references. To build the feature vectors in this case, we use a normalized count of words separated by any of the mentioned categories. This feature is only applicable to English data.

\noindent\textbf{Gender Probability: }Following the approach in \newcite{waseem2016you} we use the Twitter based lexicon presented in \newcite{sap2014developing} to calculate the probability of gender. We also convert these probabilities to binary gender by considering the positive cases as female and the rest as male. 
We make the feature vectors with the probability of the gender and binary gender for each message. This feature is not applicable to Hindi corpus.


\section{Experiments and Results}
\label{sec:results}

\subsection{Experimental Settings}
For both datasets, we trained several classification models using different combinations of features discussed in~\ref{sec:feautures}. Since this is a multi-class classification task, we use a one-versus-rest classifier which trains a separate classifier for each class and labels each comment with the class with highest predicted probability across all classifiers. We tried Logistic Regression and linear SVM as the estimator for the classifier. We decided to use Logistic Regression in our final systems, since it works better in the validation phase. We implemented all models using scikit-learn tool\footnote{\url{http://scikit-learn.org/stable/}}.

\subsection{Results}

To build our best systems for both English and Hindi data, we experimented with several models using the different combinations of available features. Table~\ref{table-dev} shows the validation results on training and validation sets.

\begin{table*}[h]
\centering

\begin{tabular}{l|l|l|}
\cline{2-3}
                                      & \multicolumn{2}{c|}{\textbf{F1-weighted}}                         \\ \hline
\multicolumn{1}{|l|}{\textbf{Feature}} & \multicolumn{1}{c|}{\textbf{English}} & \multicolumn{1}{c|}{\textbf{Hindi}} \\ \hline
\multicolumn{1}{|l|}{Unigram (U)} & 0.5804                           & 0.6159                           \\ 
\multicolumn{1}{|l|}{Bigram (B)}        & 0.4637                           & 0.5195                         \\
\multicolumn{1}{|l|}{Trigram (T)}        & 0.3846                           & 0.4300                           \\ \hline
\multicolumn{1}{|l|}{Char 3gram (C3)} & 0.5694                           & 0.6065                        \\ 
\multicolumn{1}{|l|}{Char 4gram (C4)}        & 0.5794                           & 0.6212                          \\
\multicolumn{1}{|l|}{Char 5gram (C5)}        & 0.5758                           & 0.6195                           \\ \hline
\multicolumn{1}{|l|}{Word Embeddings (W2V)}        & 0.5463                           & N/A                           \\ \hline
\multicolumn{1}{|l|}{Sentiment (S)}        & 0.3961                           & N/A                          \\ \hline
\multicolumn{1}{|l|}{LIWC}        & 0.4350                           & N/A                           \\ \hline
\multicolumn{1}{|l|}{Gender Probability (GP)}        & 0.3440                           & N/A                         \\ \hline
\multicolumn{1}{|l|}{BU + U + C4 + C5 + W2V}        & \textbf{0.5875}                           & N/A                         \\ 
\multicolumn{1}{|l|}{C3 + C4 + C5}        & 0.5494                           & 0.6207                       \\ 
\multicolumn{1}{|l|}{U + C3 + C4 + C5}        & 0.5541                           & \textbf{0.6267 }                       \\ \hline
\end{tabular}
\caption{Validation results for different features for the English and Hindi datasets using Logistic Regression model. In this table BU stands for Binary Unigram.}
\label{table-dev}
\end{table*}

Table~\ref{table2} shows the results of our three submitted systems for the English Facebook and Social Media data. In all three systems, we used the same set of features as follows: binary unigram, word unigram, character n-grams of length 4 and 5, and word embeddings. In the first system, we used both train and validation sets for training our ensemble classifier. In the second system we only used the train data for training the model. The only difference between the second and the third models is that we corrected the misspellings using PyEnchant\footnote{\url{https://pypi.org/project/pyenchant}} spell checking tool. Unfortunately, we could not try applying the sentiment and lexicon-based features after spell correction due to the restrictions on the total number of submissions. However, we believe that it can improve the performance of the system.  

\begin{table*}[h!]
\centering

\begin{tabular}{l|l|l|}
\cline{2-3}
                                      & \multicolumn{2}{c|}{\textbf{F1 (weighted)}}                         \\ \hline
\multicolumn{1}{|l|}{\textbf{System}} & \multicolumn{1}{c|}{\textbf{FB}} & \multicolumn{1}{c|}{\textbf{SM}} \\ \hline
\multicolumn{1}{|l|}{Random Baseline} & 0.3535                           & 0.3477                           \\ \hline
\multicolumn{1}{|l|}{System 1}        & 0.5673                           & 0.5453                           \\
\multicolumn{1}{|l|}{System 2}        & 0.5847                           & 0.5391                           \\
\multicolumn{1}{|l|}{System 3}        & \textbf{0.5921}                  & \textbf{0.5663}                  \\ \hline
\end{tabular}
\caption{Results for the English test set. FB: Facebook and SM: Social Media.}
\label{table2}
\end{table*}
\begin{table*}[h!]
\centering

\begin{tabular}{l|l|l|}
\cline{2-3}
                                      & \multicolumn{2}{c|}{\textbf{F1 (weighted)}}                         \\ \hline
\multicolumn{1}{|l|}{\textbf{System}} & \multicolumn{1}{c|}{\textbf{FB}} & \multicolumn{1}{c|}{\textbf{SM}} \\ \hline
\multicolumn{1}{|l|}{Random Baseline} & 0.3571                           & 0.3206                           \\ \hline
\multicolumn{1}{|l|}{System 1}        & \textbf{0.6451}                           & \textbf{0.4853}                           \\
\multicolumn{1}{|l|}{System 2}        & 0.6292                           & 0.4689                           \\ \hline
\end{tabular}
\caption{Results for the Hindi test set. FB: Facebook and SM: Social Media.}
\label{table3}
\end{table*}

Table~\ref{table3} shows the performance of our systems for the Hindi Facebook and social media data. For the Hindi dataset, the combination of word unigrams, character n-grams of length 3, 4 and 5 gives the best performance over the validation set. These features capture the word usage distribution across classes. Both System 1 and System 2 use these features, trained over training set only and training and validation sets respectively.

\subsection{Analysis}

Looking at the mislabeled instances at validation phase, we found that there are two main reasons for the classifier mistakes:
\begin{enumerate}
\item Perceived level of aggression can be subjective. There are some examples in the validation dataset where the label is CAG but it is more likely to be OAG and vice versa. Table~\ref{table5} shows some of these examples.
\item There are several typos and misspellings in the data that affect the performance.

\end{enumerate}

\begin{table}[h!]
\footnotesize
\begin{tabular}{ll|l|l|}
\cline{3-4}
                                                        &                                                                                                                                                                                                                                    & \multicolumn{2}{c|}{\textbf{Label}}  \\ \hline
\multicolumn{1}{|l|}{\textbf{Language}}                 & \textbf{Example}                                                                                                                                                                                                                   & \textbf{Actual} & \textbf{Predicted} \\ \hline
\multicolumn{1}{|l|}{\multirow{1}{*}{\textbf{English}}} & \begin{tabular}[c]{@{}l@{}}What has so far Mr.Yechuri done for this Country. Ask him to shut down his bloody\\ piehole for good or I if given the chance will crap on his mouth hole.\end{tabular}                                 & CAG             & OAG                \\ \cline{2-4} 
\multicolumn{1}{|l|}{}                                  & \begin{tabular}[c]{@{}l@{}}The time you tweeted is around 3 am morning,,which is not at all a namaz time.,As\\ you bollywood carrier is almost finished, you are preparing yourself for politics by\\ these comments.\end{tabular} & OAG             & CAG                \\ \hline
\multicolumn{1}{|l|}{\multirow{2}{*}{\textbf{Hindi}}}   & ajeeb chutya hai.... kahi se course kiya hai ya paida hee chutya hua tha                                                                                                                                                           & CAG             & OAG                \\ \cline{2-4} 
\multicolumn{1}{|l|}{}                                  & \begin{tabular}[c]{@{}l@{}}Salman aur aamir ki kounsi movie release huyee jo aandhi me dub gaye?? ?Bikau\\ chatukar media\end{tabular}                                                                                             & OAG             & CAG                \\ \hline
\end{tabular}
\caption{Misclassified examples in case of the aggression level}\label{table5}
\end{table}

\normalsize

\begin{figure*}[h!]
\begin{subfigure}{.5\textwidth}
  \centering
\includegraphics[width=0.90\textwidth, height=0.20\textheight]{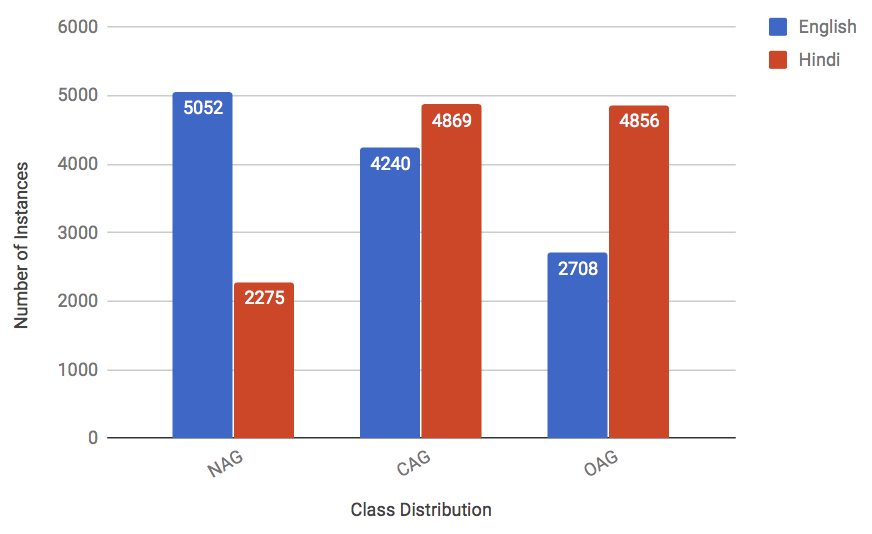}
\caption{Data distribution for training sets}
\label{fig:1data}
\end{subfigure}%
\begin{subfigure}{.5\textwidth}
  \centering
\includegraphics[width=0.90\textwidth, height=0.20\textheight]{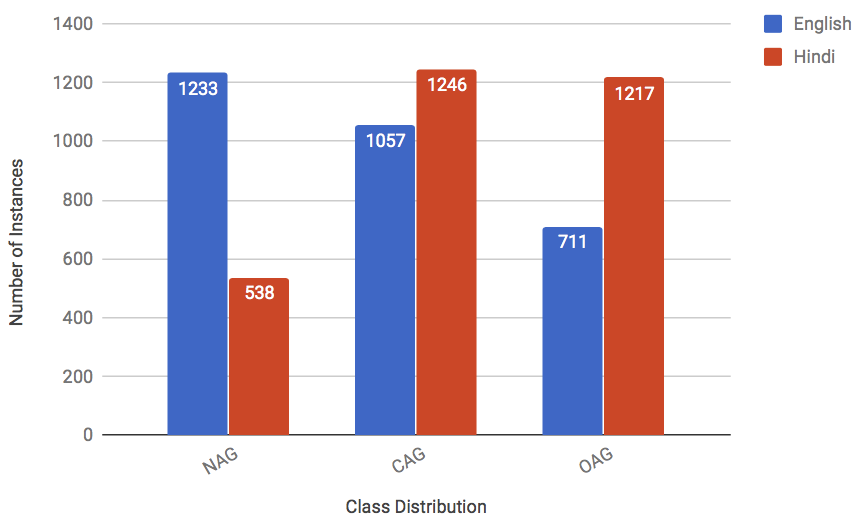}
\caption{Data distribution for evaluation sets}
\label{fig:2data}
\end{subfigure}
\caption{Label distribution comparison between training and evaluation sets}
\label{fig:datafig}
\end{figure*}

Also, it is obvious from Figure~\ref{fig:datafig} that Hindi corpus is more balanced than the English one in case of OAG and CAG instances. That could be a good reason why the performance of the lexical features is better for Hindi data.

Table \ref{top_n_hin} illustrates the most informative features learned by the classifier for all three classes in Hindi data. We observe that word unigrams and character trigrams are the most important features for the system. From the table, the top features for CAG are mostly swear words in Hindi and character n-grams of the swear words. More English words appear in the top list for NAG than the other two classes. There is no overlap between these features with top features from either CAG or OAG. Our system has difficulty differentiating between OAG and CAG when there is no strong swear word in the comments. 


\begin{table*}[h!]
\centering

\begin{tabular}{|c|c|c|}
\hline
\textbf{NAG} & \textbf{CAG} & \textbf{OAG} \\
\hline
unigram\_: &unigram\_?&char\_tri\_gram\_kut\\
unigram\_mera &unigram\_baki&unigram\_bc\\
unigram\_bike  &unigram\_??&char\_tri\_gram\_cho\\
unigram\_jai   &unigram\_pm&char\_4\_gram\_ kut\\
unigram\_main  &unigram\_o&unigram\_chutiya\\
unigram\_sahi  &char\_tri\_gram\_ ky&unigram\_maa\\
unigram\_........... &unigram\_badla&unigram\_mc\\
unigram\_launch &cha\_tri\_gram\_yad&unigram\_gand \\
unigram\_jay   &char\_5\_gram\_e...&char\_tri\_gram\_tiy\\ 
char\_tri\_gram\_mer &unigram\_3&char\_tri\_gram\_chu\\  
\hline

\end{tabular}
\caption{Top 10 features learned by System 1 for each class for the Hindi dataset.}
\label{top_n_hin}
\end{table*}

\begin{figure*}[h!]
\begin{subfigure}{.5\textwidth}
  \centering
\includegraphics[width=0.85\textwidth, height=0.25\textheight]{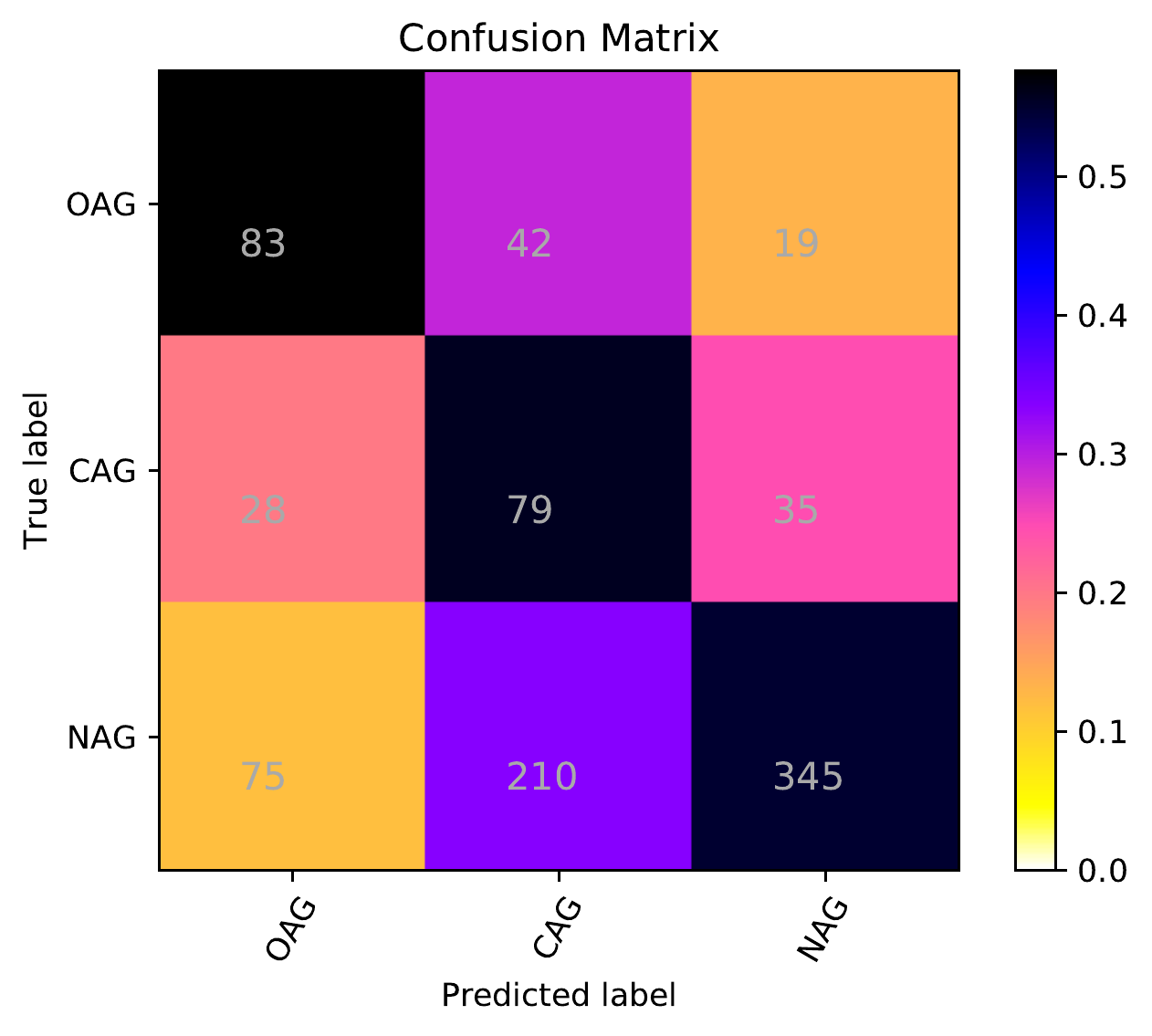}
\caption{EN-FB task}
\label{fig:1}
\end{subfigure}%
\begin{subfigure}{.5\textwidth}
  \centering
\includegraphics[width=0.85\textwidth, height=0.25\textheight]{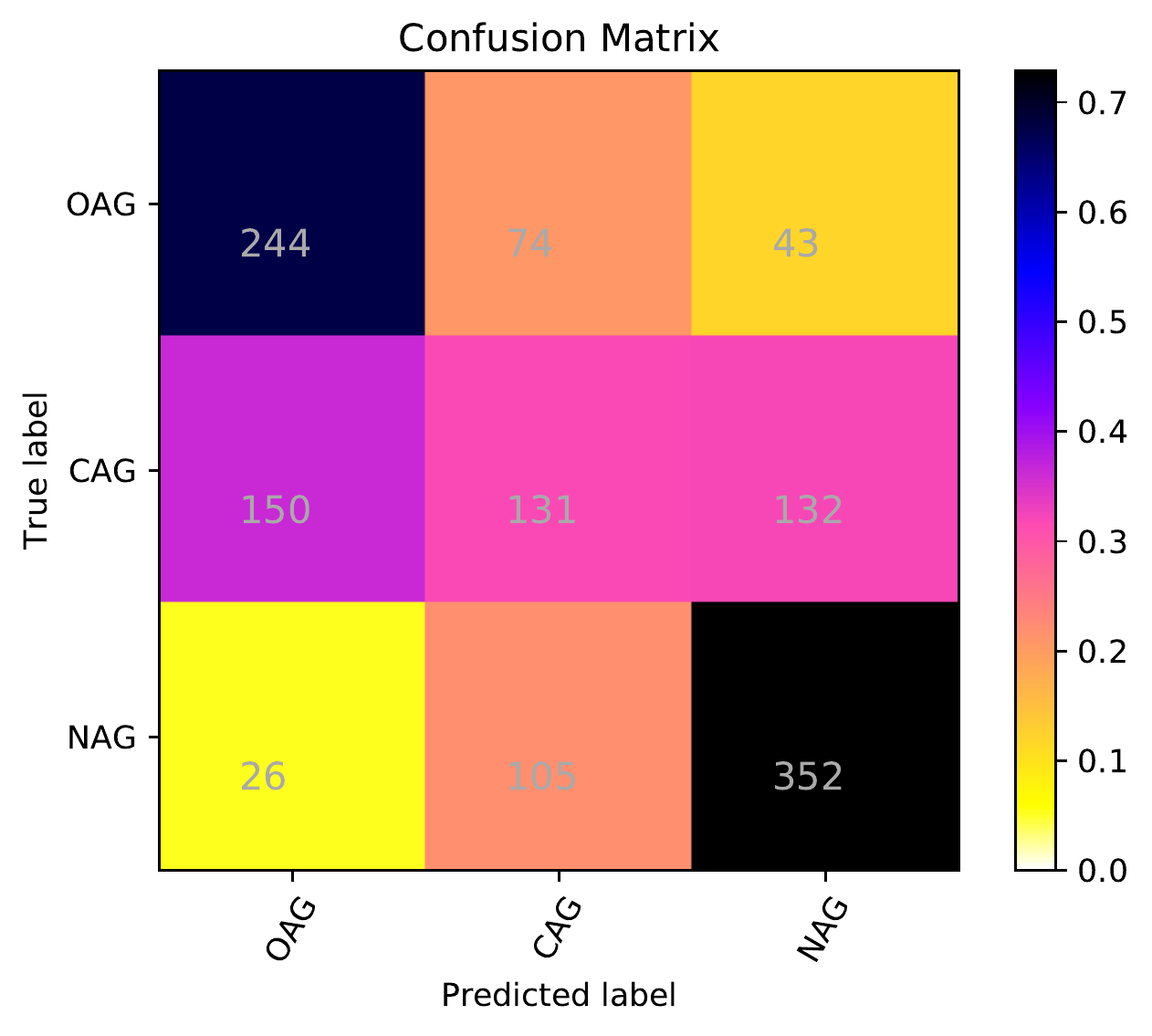}
\caption{EN-SM task}
\label{fig:2}
\end{subfigure}
\caption{plots of confusion matrices of our best performing systems for English Facebook and Social Media data}
\label{fig:fig}
\end{figure*}

Figure~\ref{fig:1} shows the confusion matrix of our best model for all three classes in English Facebook corpus. The most interesting part of this figure is that the classifier mislabeled several NAG instances with CAG label. Since our system is mostly based on lexical features, we can conclude that there are much fewer profanities in CAG instances comparing with the OAG ones, which make it hard to distinguish them from NAG examples without considering the sentiment aspects of the messages. This fact can also be proved by looking at Figure~\ref{fig:2}, since it seems that the classifier was also confused to label CAG instances in both cases with and without profanities in English Social Media corpus.

\begin{figure*}[h!]
\begin{subfigure}{.5\textwidth}
  \centering
\includegraphics[width=0.85\textwidth, height=0.25\textheight]{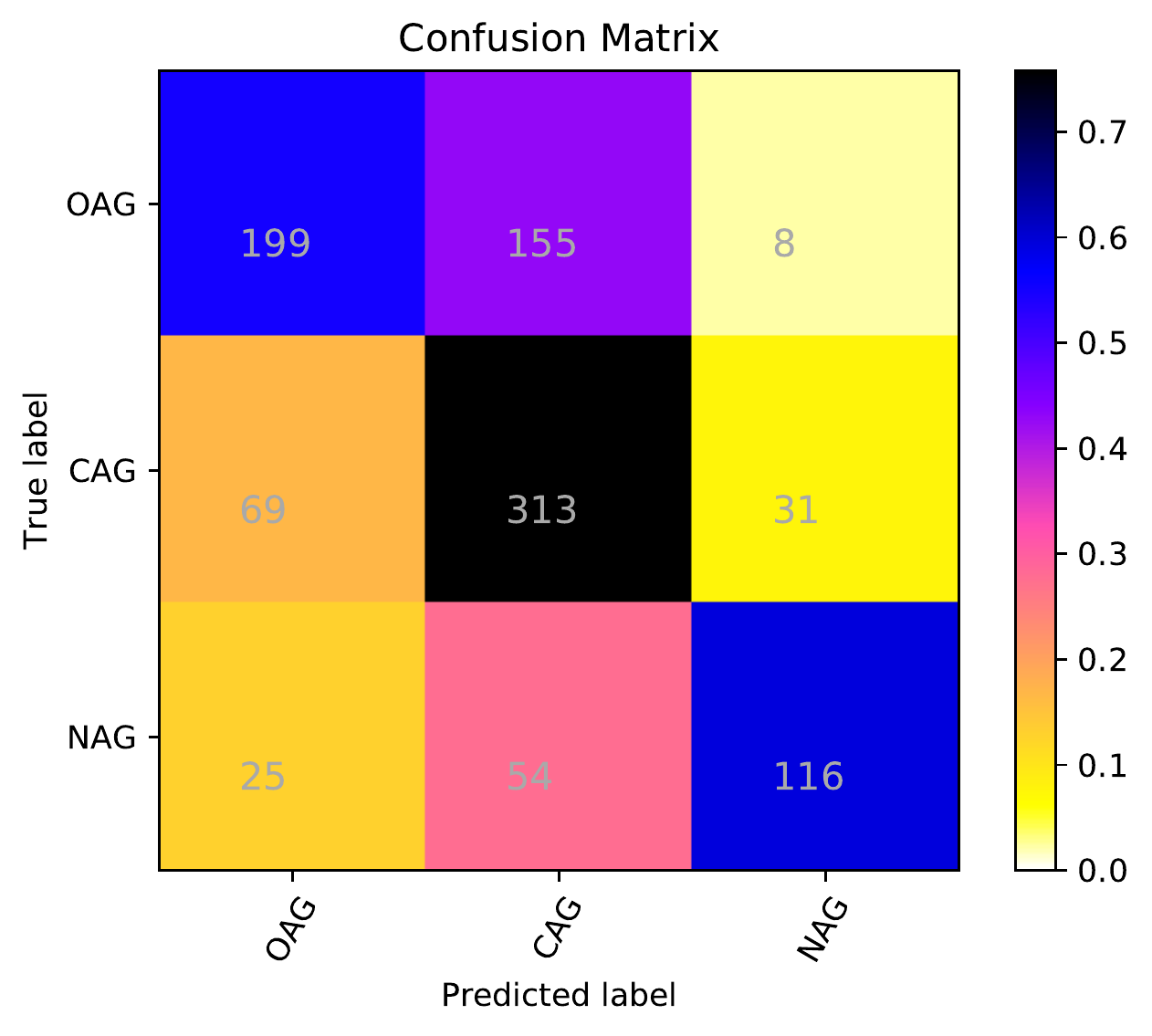}
\caption{HI-FB task}
\label{fig:3}
\end{subfigure}%
\begin{subfigure}{.5\textwidth}
  \centering
\includegraphics[width=0.85\textwidth, height=0.25\textheight]{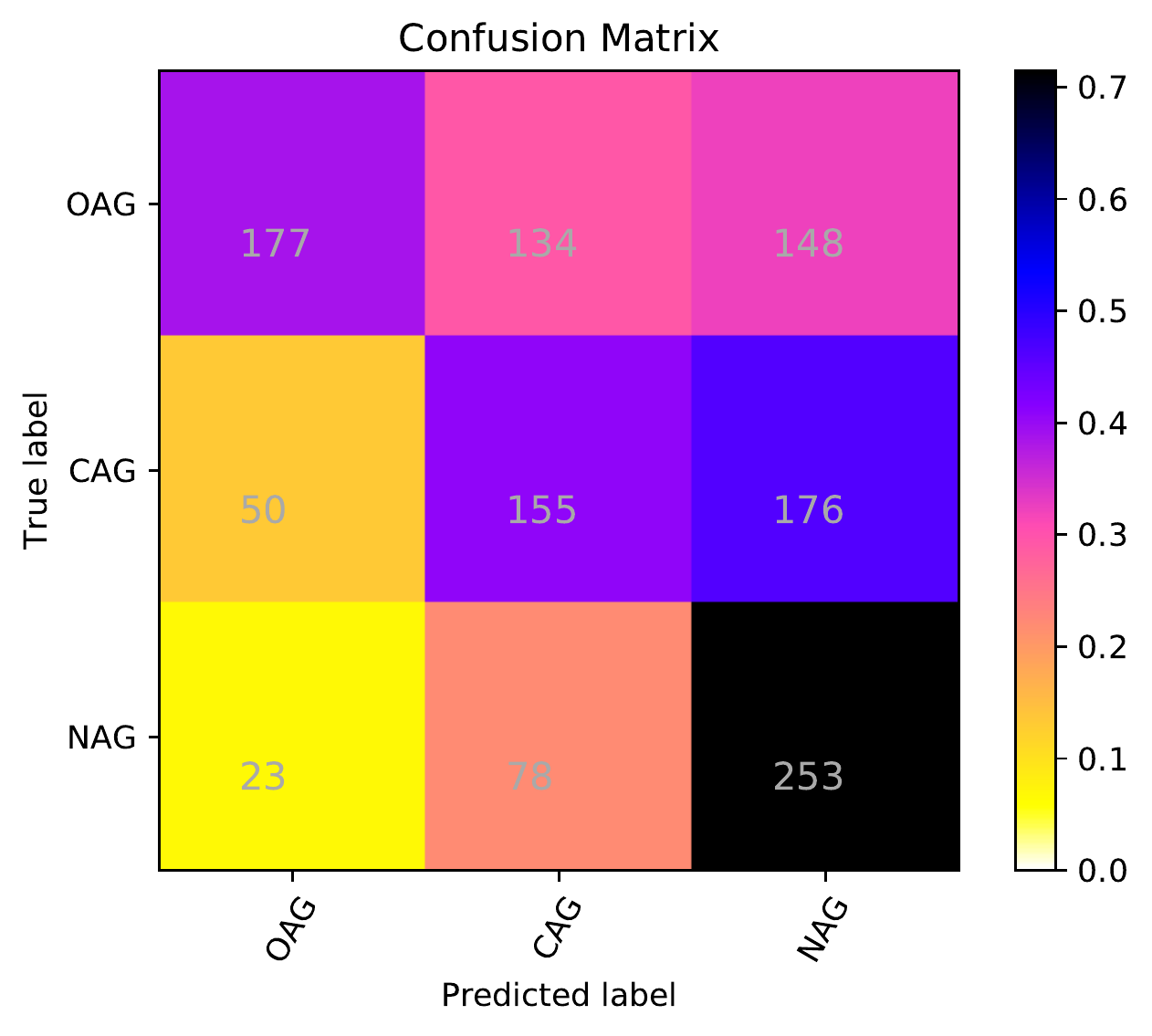}
\caption{HI-SM task}
\label{fig:4}
\end{subfigure}
\caption{Plots of confusion matrices of our best performing systems for Hindi Facebook and Social Media data}
\label{fig:fig*}
\end{figure*}

Figure~\ref{fig:3} shows that for Hindi Facebook data, the most biggest challenge is to distinguish OAG instances from CAG ones. Since our proposed system, in this case, is completely built on lexical features, it can be inferred from the figure that even indirect aggressive comments in Hindi language contains lots of profanities. However, for the Hindi Social Media corpus, we have the same concern as English data.  

\section{Conclusion}

In this paper, we present our approaches to identify the aggression level in English and Hindi comments in two different datasets, one from Facebook and another from other social media. In our best performing systems, we use a combination of lexical and semantic features for English corpus, and lexical features for Hindi data. 

Future work for English data includes exploring more sentiment features to capture implicit hateful comments and adding more pre-processing levels. For instance, non-English character removal can improve the system since our proposed model is mainly based on lexical features, and is likely very sensitive to unknown characters and words. For the Hindi dataset, identifying the Hindi-English code-mixed instances and processing these instances and Hindi monolingual instances separately could be a future direction to explore. As the classification of aggression is subjective in most scenarios, adding sentiment features to the lexical information might help to model performance for Hindi data. 

\color{black}
\bibliography{trac}

\begin{thebibliography}{}

\bibitem[\protect\citename{Davidson \bgroup et al.\egroup
  }2017]{davidson2017automated}
Thomas Davidson, Dana Warmsley, Michael Macy, and Ingmar Weber.
\newblock 2017.
\newblock {Automated Hate Speech Detection and the Problem of Offensive
  Language}.
\newblock In {\em Proceedings of ICWSM}.

\bibitem[\protect\citename{Kumar \bgroup et al.\egroup }2018a]{trac2018report}
Ritesh Kumar, Atul~Kr. Ojha, Shervin Malmasi, and Marcos Zampieri.
\newblock 2018a.
\newblock {Benchmarking Aggression Identification in Social Media}.
\newblock In {\em Proceedings of the First Workshop on Trolling, Aggression and
  Cyberbulling (TRAC)}, Santa Fe, USA.

\bibitem[\protect\citename{Kumar \bgroup et al.\egroup }2018b]{trac2018dataset}
Ritesh Kumar, Aishwarya~N. Reganti, Akshit Bhatia, and Tushar Maheshwari.
\newblock 2018b.
\newblock {Aggression-annotated Corpus of Hindi-English Code-mixed Data}.
\newblock In {\em Proceedings of the 11th Language Resources and Evaluation
  Conference (LREC)}, Miyazaki, Japan.

\bibitem[\protect\citename{Le and Mikolov}2014]{le2014distributed}
Quoc~V. Le and Tomas Mikolov.
\newblock 2014.
\newblock Distributed representations of sentences and documents.
\newblock In {\em ICML}, volume~14, pages 1188--1196.

\bibitem[\protect\citename{Malmasi and Zampieri}2018]{malmasi2018challenges}
Shervin Malmasi and Marcos Zampieri.
\newblock 2018.
\newblock {Challenges in Discriminating Profanity from Hate Speech}.
\newblock {\em Journal of Experimental \& Theoretical Artificial Intelligence},
  30:1--16.

\bibitem[\protect\citename{Mave \bgroup et al.\egroup }2018]{mave2018lid}
Deepthi Mave, Suraj Maharjan, and Thamar Solorio.
\newblock 2018.
\newblock {Language Identification and Analysis of Code-Switched Social Media
  Text}.
\newblock In {\em Proceedings of the Third Workshop on Computational Approaches
  to Linguistic Code-Switching}, Melbourne, Australia, July. Association for
  Computational Linguistics.

\bibitem[\protect\citename{Nobata \bgroup et al.\egroup
  }2016]{nobata2016abusive}
Chikashi Nobata, Joel Tetreault, Achint Thomas, Yashar Mehdad, and Yi~Chang.
\newblock 2016.
\newblock {Abusive Language Detection in Online User Content}.
\newblock In {\em Proceedings of the 25th International Conference on World
  Wide Web}, pages 145--153. International World Wide Web Conferences Steering
  Committee.

\bibitem[\protect\citename{Pennebaker \bgroup et al.\egroup
  }2007]{Pennebaker07liwc}
James~W. Pennebaker, Roger~J. Booth, and Martha~E. Francis.
\newblock 2007.
\newblock Liwc2007: Linguistic inquiry and word count.
\newblock {\em Austin, Texas: liwc.net}.

\bibitem[\protect\citename{Samghabadi \bgroup et al.\egroup
  }2017]{samghabadi2017detecting}
Niloofar~Safi Samghabadi, Suraj Maharjan, Alan Sprague, Raquel Diaz-Sprague,
  and Thamar Solorio.
\newblock 2017.
\newblock Detecting nastiness in social media.
\newblock In {\em Proceedings of the First Workshop on Abusive Language
  Online}, pages 63--72.

\bibitem[\protect\citename{Sap \bgroup et al.\egroup }2014]{sap2014developing}
Maarten Sap, Gregory Park, Johannes Eichstaedt, Margaret Kern, David Stillwell,
  Michal Kosinski, Lyle Ungar, and Hansen~Andrew Schwartz.
\newblock 2014.
\newblock Developing age and gender predictive lexica over social media.
\newblock In {\em Proceedings of the 2014 Conference on Empirical Methods in
  Natural Language Processing (EMNLP)}, pages 1146--1151.

\bibitem[\protect\citename{Socher \bgroup et al.\egroup
  }2013]{socher2013recursive}
Richard Socher, Alex Perelygin, Jean Wu, Jason Chuang, Christopher~D Manning,
  Andrew Ng, and Christopher Potts.
\newblock 2013.
\newblock Recursive deep models for semantic compositionality over a sentiment
  treebank.
\newblock In {\em Proceedings of the 2013 conference on empirical methods in
  natural language processing}, pages 1631--1642.

\bibitem[\protect\citename{Van~Hee \bgroup et al.\egroup }2015]{VanHee15detect}
Cynthia Van~Hee, Els Lefever, Ben Verhoeven, Julie Mennes, Bart Desmet, Guy
  De~Pauw, Walter Daelemans, and Veronique Hoste.
\newblock 2015.
\newblock Detection and fine-grained classification of cyberbullying events.
\newblock In {\em Proceedings of the International Conference Recent Advances
  in Natural Language Processing}, pages 672--680. INCOMA Ltd. Shoumen,
  Bulgaria.

\bibitem[\protect\citename{Waseem and Hovy}2016]{waseem2016hateful}
Zeerak Waseem and Dirk Hovy.
\newblock 2016.
\newblock Hateful symbols or hateful people? predictive features for hate
  speech detection on twitter.
\newblock In {\em Proceedings of the NAACL student research workshop}, pages
  88--93.

\bibitem[\protect\citename{Waseem}2016]{waseem2016you}
Zeerak Waseem.
\newblock 2016.
\newblock Are you a racist or am i seeing things? annotator influence on hate
  speech detection on twitter.
\newblock In {\em Proceedings of the first workshop on NLP and computational
  social science}, pages 138--142.

\bibitem[\protect\citename{Wulczyn \bgroup et al.\egroup }2016]{Wulczyn:2016}
Ellery Wulczyn, Nithum Thain, and Lucas Dixon.
\newblock 2016.
\newblock Ex machina: Personal attacks seen at scale.
\newblock {\em CoRR}, abs/1610.08914.

\end{thebibliography}
\bibliographystyle{acl}

\end{document}